# Event-Driven Imaging in Turbid Media: A Confluence of Optoelectronics and Neuromorphic Computation

NING ZHANG,[1,*] TIMOTHY SHEA,[2] AND ARTO NURMIKKO [1,*]

[1]*School of Engineering, Brown University, 184 Hope St, Providence, RI, 02912, USA*
[2]*Intel Laboratories, Sunnyvale, CA, 94086, USA*
*\*ning_zhang1@brown.edu, arto_nurmikko@brown.edu*

**Abstract:** In this paper a new optical-computational method is introduced to unveil images of targets whose visibility is severely obscured by light scattering in dense, turbid media. The targets of interest are taken to be dynamic in that their optical properties are time-varying whether stationary in space or moving. The scheme, to our knowledge the first of its kind, is human vision inspired whereby diffuse photons collected from the turbid medium are first transformed to spike trains by a dynamic vision sensor as in the retina, and image reconstruction is then performed by a neuromorphic computing approach mimicking the brain. We combine benchtop experimental data in both reflection (backscattering) and transmission geometries with support from physics-based simulations to develop a neuromorphic computational model and then apply this for image reconstruction of different MNIST characters and image sets by a dedicated deep spiking neural network algorithm. Image reconstruction is achieved under conditions of turbidity where an original image is unintelligible to the human eye or a digital video camera, yet clearly and quantifiable identifiable when using the new neuromorphic computational approach.

## 1. Introduction

The presence of dense light-scattering opaque media, whether atmospheric fog, biological tissue, underwater turbidity, or other heterogeneous ambient presents a challenge to the identification and imaging of dynamical targets of interest across a wide range of applications such as in medical imaging, tracking moving objects, or free-space transmission of structured light. Many approaches have been reported over the years combining optical engineering with specific model image reconstruction algorithms which are usually focused on specific circumstances such as encountered in bioimaging through body tissue by diffusive optical tomography (DOT) [2], in the detection of vehicles moving in a fog [4], identification of objects in turbid underwater media [5], to cite but a couple of examples from the literature on the subject. Advances to solving the challenges have been made with the help of a new generation of optical engineering tools such as time-gated and related time-of-flight (ToF) detection techniques [6] and the availability of single-photon counting detector arrays [7] on one hand, as well as the recent adaptation of powerful deep machine-learning computational techniques on the other [8]. Yet, the spatio-temporal resolution achievable by present state-of-the-art techniques for image reconstruction remains limited when the optical thickness of the turbid media exceeds a few tens of photon mean free paths (MFP). The difficulties are related irrespective of whether an optical system operates in backscattering mode such as in LIDAR [9], or in transmission mode such as in fluorescence microscopy [10] or line-of-sight communication [11].

In this paper we present an entirely new approach to optical imaging in dense turbid media (MFP> 50) by pairing a dynamic vision sensor (DVS) [12] with neuromorphic computing, the latter implemented through a deep spiking neural network approach (SNN) [13]. We focus specifically on the problem of imaging in dense turbid media when a target of interest is dynamic because its optical appearance is time varying or it is in kinematic motion, or some combination of the two. The turbid media itself is taken as static or slowly varying with scattering as the dominant optical loss mechanism, quantified by $\mu_s$ the scattering coefficient, and g the anisotropy factor [14]. Given the intrinsic event-detecting feature of an DVS-type camera in responding only to time-varying illumination, this property helps to suppress the generally dominant static or quasi-static background in scattered light emerging from any turbid medium. Our approach is inspired by the exquisite sensitivity of human vision in event detection, that is the synergy of the retina with the brain where neuronal spikes are the one global currency of information. Here the retina is approximated by the DVS camera which only captures time-varying events and converts these into trains of impulses akin to spikes in asynchronous neuronal firing, at each pixel. The spike train outputs from the pixels of the camera are forwarded to a neuromorphic computational engine acting as an analog to the human brain engaged in pattern recognition. Exploiting recent advances in neuromorphic computing we develop a physics-based customized deep SNN algorithm which decodes the camera-delivered spike-based data and, once trained, is capable of image reconstruction in real time.

Below we present results which demonstrate neuromorphic-based optical imaging to unveil dynamical objects of complex shape in a regime where normal imaging techniques are challenged in their ultimate spatial resolution. For benchtop simplicity, we focus on the case where the targets themselves are static in space but their optical appearance is dynamic. In a co-designed approach, optical experiments using a near infrared laser and a DVS camera in either reflection (backscattering) or transmission geometry are combined with a neuromorphic computing approach for which we developed a computationally efficient deep-learning SNN algorithm. We first describe the event-detecting optical system using a well-characterized phantom as the proxy for dense light-scattering medium. This is followed by a short introduction to neuromorphic computing before the description of our deep SNN model and the method for training the SNN protocol. The performance of the full end-to-end system is demonstrated through examples of image reconstruction using three widely used character and image sets as targets, namely the Kanji-MNIST, Fashion-MNIST and MNIST sets. The results show how our technique is able to recover clearly identifiable, quantitatively measurable images after light propagation through diffusive media when either a human eye or a digital camera is unable to do so.

## 2. Results

**Event-driven optical system**

Our objective is to develop a generalizable proof-of-concept for image reconstruction of a dynamical object, both spatially and temporally, that is embedded within a diffusive medium. For experimental simplicity we used spatially stationary targets endowed with programmable dynamical optical images as targets. However, the method described in this paper is also applicable to spatially moving targets. Figure 1 illustrates the experimental setup of the Event-Driven Neuromorphic Optical Subsystem (EDOS) with a continuous wave near-infrared laser as the source (850 nm, 15 mW). While configured on benchtop in the laboratory, the scheme can be also implemented for field portable or body wearable use by using compact components such as a low-power semiconductor laser, miniaturized optical lenses and so on. Here, after collimated beam expansion the beam spot had a radius of 15 mm with a corresponding illumination fluence of 0.021mW mm$^{-2}$. A 50-50 beam splitter was used to divide the expanded beam into two paths, facilitating experiments both in reflection and transmission geometry, respectively, according to the optical paths shown in upper and lower halves of Fig 1.

In reflection geometry we placed the target behind a densely scattering medium, to mimic a case such as in LIDAR sensor applications that might involve the identification of a vehicle or pedestrian obscured by fog, or hemodynamically active vasculature in the brain in biomedical imaging of cortical metabolism. In each instance, a source and a detector locate on the same plane so that photons launched from the source must traverse the scattering medium twice before backscattered light is collected at the detector. Among many possible choices for proxy scattering media [15] we constructed silicone-based phantoms with microscale $SiO_2$ particles as scatterers (see Methods). In the experiments, the scattering coefficient of the phantom was $\mu_s$ = 6 mm$^{-1}$, anisotropy number g is 0.9 [16] so that for a slab thickness of 5 mm the resulting number of mean free paths was MFP = 60 in the reflection geometry.

To simulate dynamic targets with many different spatial features we used an electronic ink (E-ink) display [17] located behind the phantom. Commercially available E-ink displays are particularly convenient as passive optical components with a programmable display where manipulation of magnetic minispheres alters the absorption properties in providing a black-an-white display content. Due to the stochastic nature of light scattering and photon detection by the event-based sensor electronics, there is no deterministic relationship between the original dynamic targets and the event sequences recorded by the DVS. For this reason, we applied two methods to generate alternative 'ground truth' data to be used when training our neuromorphic deep learning model.

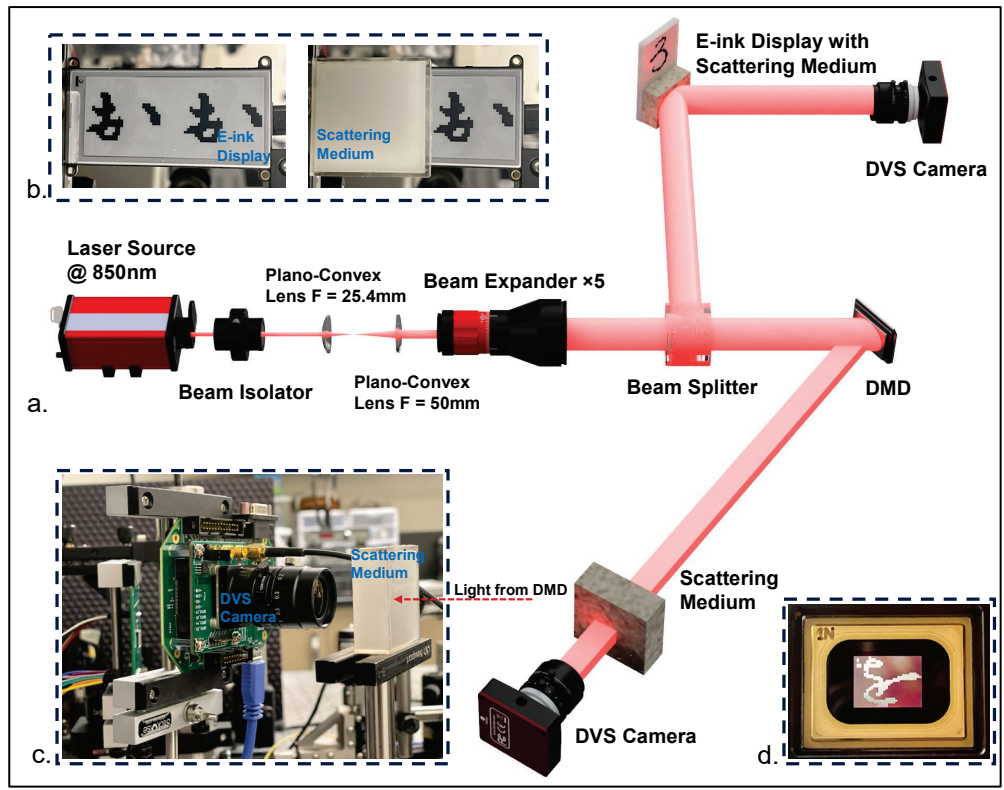

Fig 1. Event-Driven Neuromorphic Optical Subsystem (EDOS) and its experimental design as composite view. (a) The beam from a low power infrared laser (λ=850nm) is first expanded, then divided by a 50-50 beamsplitter for experiments in reflection and transmission geometry, respectively. The "upper" beam in the figure applies to reflection experiments where a silicone-based optically thick scatterer (a phantom with photon MFP = 60) is positioned in front of an E-ink display. The display is split-screen programmed to present a pair of duplicate images (400 ms "flash" of each target), one being obscured by the phantom and the other without any impediment to free space propagation. A DVS camera captures reflected photons from both targets simultaneously. In the transmission experiments, the "lower beam" illuminates a Digital Micromirror Device (DMD) which projects programmed sets of images onto the scattering phantom (here a 1ms flash of each target, MFP = 72) with the DVS camera now behind the phantom to capture transmitted photons. (b) Illustration of one Kanji character duplicate pattern generated by the E-ink display (left image), and room-light view of the reflected light with and without a phantom (right image). (c) A photo from the laboratory benchtop showing the positioning of the DVS camera and a phantom (MFP = 72) in transmission experiments. (d) Example of an image of the DMD projection of one Kanji character.

First, we programmed the E-ink device to show simultaneously two copies of a given target content whereby the bottom image was hidden behind the phantom and completely obscured to the naked eye while the unobscured top image was used to provide an experimental reference as ground truth. This serves to generate a valid and statistically similar spike sequence to what the DVS would emit in the absence of the phantom, though precise spike times and coordinates will not be the identical. Simultaneous dual display of both copies also allowed for convenient time synchronization of the DVS camera in a protocol where the images were flashed periodically on and off, typically at a rate of ~ 2.5Hz (refresh speed limit of a common E-ink display) to mimic dynamical targets. Second, when the output of the last layer in the SNN model is a time-integrated membrane potential generated by accumulating the spiking activity, the ground truth can be simply taken as digital images from libraries directly. In this scenario, the reconstruction only focuses on spatial geometry.

In the transmission experiments (lower portion of *Fig. 1*) photons from the laser source traveled through the phantom before detection with the dominant losses again obscuring any spatial features by scattering. Here we used a digital mirror device (DMD, Vialux Inc) to project structured light periodically (up to 1000 Hz) in the form of various geometrical shapes directly onto the surface of another slab of the silicone-based scattering medium, now with a thickness of 12 mm corresponding to optical thickness of 72 MFPs (with $\mu_s$ = 6 mm$^{-1}$ and g=0.9). The DVS camera collected the transmitted photons, again tasked with detecting and converting time-varying signals into asynchronous trains of spikes. The ground truth spike trains were collected in experiments in absent of the scattering media.

For both the reflection and transmission experiments we used a commercial DVS camera by Prophesee Inc (Gen3.0 EVK) as the detector. Equipped with an active area of 9.6 × 7.2 mm2 with 640 × 480 pixels the camera includes a lens system mounted in front of the CMOS sensor chip to enhance the collection efficiency of photons. As noted, a DVS camera operates by detecting and responding to changes in light intensity at each pixel, in real-time. Unlike traditional frame-based cameras that capture a series of frames at a fixed rate, the DVS generates a spike train only when the relative light intensity at a specific pixel is subject to a dynamical change. This key feature makes the device a highly efficient event-driven sensor with a high dynamic range of up to 140 dB [18] and a high temporal resolution in the order of microseconds (each pixel), with low energy consumption and minimal latency, significant assets for any real-time optical imaging system.

The overall optical configuration of Fig 1 is relatively straightforward when compared with typical configurations of other advanced diffusive imaging techniques for dynamical targets whether based on Optical Coherence Tomography (OCT) [19] or Diffuse Optical Tomography (DOT) [20]. When combined with chip scale semiconductor laser sources, our relatively simple architecture suggests future potential for integrating the system as one monolithic ultracompact optical device in the reflection mode.

**Neuromorphic computational imaging (NCI) - Rationale**

In the growing field of field of neuromorphic computing and engineering [21,22], deep learning approaches have shown their utility when dealing with high-dimensional problems associated with computer vision, for example to enhance robotic navigation in crowded environments [23]. For the work in this paper we developed a special purpose neuromorphic model using deep spiking neural network (SNN) surrogate gradient training algorithms [24]. While brain inspired, these algorithms leverage robust auto-differentiation frameworks (e.g. PyTorch [25]) to fit potentially millions of parameters to a training dataset. This is a logical choice given our goal of preserving the event-driven spikes received from the DVS as the principal currency of information throughout the imaging system for reconstructing target images obscured by dense turbid media. From the viewpoint of minimizing system energy consumption and reconstruction latency, computing with spikes offers advantages derived from the intrinsic spatiotemporal sparsity. In both brain circuits and current deep learning accelerators (e.g. CPUs, GPUs), the most energy-intensive operation is the transmission of signals from pre-synaptic neurons to post-synaptic neurons – achieved through axonal propagation and matrix-matrix multiplication, respectively. Biological neurons limit this expensive communication by acting as temporal integrators until activation exceeds a biophysical threshold. This feature is emulated in emerging neuromorphic hardware [26], where subthreshold neurons generate no computation related to event transmission. The benefits of this approach are greatest when the data being processed are sparse vectors or tensors abundant in zeroes, as in the case of DVS. Since neuron inputs are multiplied with synaptic weights and accumulated in the neuron state, it follows that if most input values are zero ('0'), most of the neurons will require no expensive communication. The argument for improved reconstruction latency is related. Traditional signal processing requires that all channels (camera pixels) comply with a global sampling rate. In most non-trivial imaging conditions, activation of channels is uneven. This requires a compromise between running the system more slowly to allow the least active channels time to accumulate versus running more quickly to reduce the delay before processing the most active channels. In contrast, in event-driven processing there is no forced global synchrony in the transmission of channels, resulting in lower latency in these types of specialized circumstances.

We show next how a properly designed neuromorphic SNN can exploit the inherent characteristics of high efficiency, temporal data processing, and adaptability to manage dynamic and spatially complex optical data collected by a DVS camera from a turbid medium. In particular, our novel algorithm integrates within the SNN an autoencoder structure while applying denoising to diffusive imaging. Figure 2 shows a high level view of our neuromorphic computational imaging (NCI) system in which neuron mimicking spike trains provide the single information currency throughout, from photonic event detection by the EDOS subsystem to the final target reconstruction by a SNN algorithm.

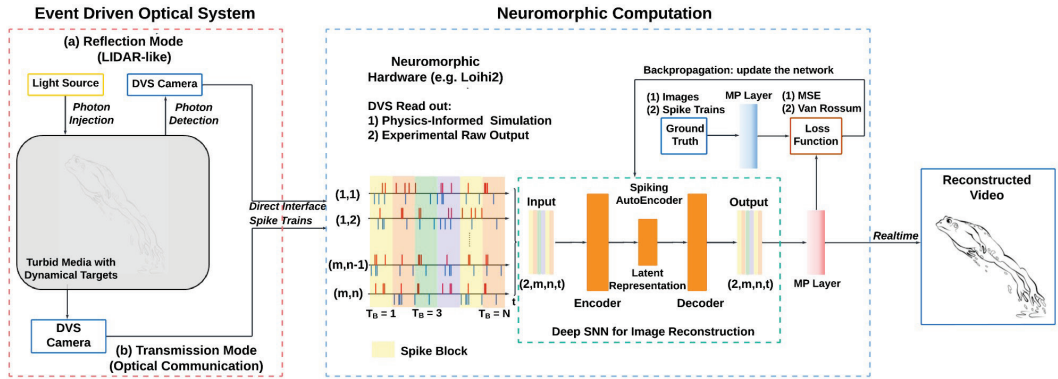

Fig 2. High level view design of the Neuromorphic Computational Imaging (NCI) scheme for reconstructing dynamical images of targets obscured by turbid media (here schematicaly a cartoon of a jumping frog). The flow of data reflects the integration of the benchtop Event-Driven Optical Subsystem (EDOS) (left) with the computational architecture of the NCI (right). Asynchronous spike trains generated by the EDOS include only those photons which have interacted with a dynamical target and provide the input to the NCI. Spikes can be also produced *in silico* from physics-based simulations such as Monte Carlo simulations. At the heart of the NCI is the deep Spiking Neural Network (SNN) engine which is trained for image reconstruction. Spikes within each timeblock are accumulated at the final membrane potential layer to form a single image frame. The actual frame rate dependings on the size of the spike blocks. For low power and minimal latency, the computing performance of the NCI can in principle be further accelerated using recently developed special purpose neuromorphic computing hardware such as the Intel's Loihi 2 neuromorphic processor [1] or IBM's True North [3].

## Design of a deep spiking neural network model for dynamical optical imaging in turbid media

The principal aim in our neuromorphic computing application for imaging in turbid media is to set up and train an SNN network, to discover those low dimensional representations of the targets that are enveloped within the scattered photons emerging from the turbid media. The box labeled 'Neuromorphic Computation' in Fig 2 shows the design of our spike-based process pipeline and key features of the spiking deep neural network, laying out the process flow and enumerating the steps involved in feature extraction. The SNN is constructed to generate a low dimensional representation of large multidimensional data in which one optimizes the weights of the network to minimize the difference (or "reconstruction error") between the original target and the reconstructed output of the SNN engine.

We approached building the SNN engine by creating a denoising Spiking Autoencoder (SAE) at its core. The SAE primarily uses clear spike trains or clear images as ground truth to compute the loss with the input spike trains. The neuron layers in the SAE are modeled as current-based leaky integrate-and-fire (LIF) neurons, a choice influenced by their biological accuracy and computational efficiency. Figure 3 shows the structure and process flow highlighting the SAE design and the role of the latent space within the SNN.

The initial inputs to the SNN are spike trains either from the physics-based in silico simulation (Monte Carlo) or from benchtop experiment DVS raw multipixel output. In this approach, the spatial-temporal coordinates of each spike are represented in an array format [p, x, y, t], where p signifies spike polarity (+1 or -1; increasing or decreasing light intensity), x and y are coordinate values of that pixel in sensor's x-y plane, t is time step in microsecond. Depending on the situational needs such as the case in this paper, spike trains may first require an event preprocessing step. The event preprocessing includes: (1) selection of data region of interest (ROI) so as to only select spike trains from a certain number of sensor pixels; (2) noise filtering (Activity-Noise Filter, Spatio-Temporal-Contrast Filter, AntiFlicker Filter) to filter out the noisy spikes out of spike trains; (3) binning process in both space and time, to bin the spike trains into a smaller number of pixel size and time steps for faster processing (see Methods).

After preprocessing, the spike trains are fed into the input layer of SAE. The SAE is composed of three main parts: the encoder, the latent space and the decoder (as in Fig 3a). The encoder's job is to convert the input data into a lower-dimensional representation, often referred to as the "latent space" or "code". This representation tries to capture the most important features of the input data. As a tutorial aid, Fig 3b illustrates the spiking activity of 128 neurons in latent space across 10 time steps acquired from the same dynamic optical target. The similarity among spike trains post-encoding and reflecting the SAE's ability to capture and extract the target's essential features. To help visualize these features in a reduced dimensionality space, t-Distributed Stochastic Neighbor Embedding (t-SNE) was employed by which the encoded spiking activity was transformed into representation in 2D space defined. Figures 3c and 3d show t-SNE cluster plots for the Kanji and Fashion-MNIST targets, respectively, each comprising 10 major classes of images. The Kanji-MNIST clusters are more distinct compared to those of Fashion-MNIST, unsurprising given their distinctive and lesser overlap of features. The t-SNE visualizations affirm qualitatively that the spiking autoencoder is effective at feature extraction, even for obscured targets. The decoder takes this representation and attempts to recreate the original unobscured targets as closely as possible. Due to the structure of the SAE, the output layer has the same number of neurons as the input layer. The output layer consists of two types of data: membrane potential and generated spikes. Both sets can be used to form the reconstructed images depending on the imaging requirements.

Both physics-based simulations and DVS experimental output are of dynamical nature and can reach high temporal resolution. We trained the SAE by individual time steps as follows. The neurons in the final layer were set to have a high threshold, sufficient for accumulating the membrane potential while minimizing spike generation randomness. The spikes inside the same spike block were processed at each time step and then accumulated at the final layer to reconstruct one image. A series of spike blocks (such as shown in Fig. 6 below) then can be reconstructed as a video for real time visualization of the target dynamics. The rate of the video depends on the size of the spike blocks.

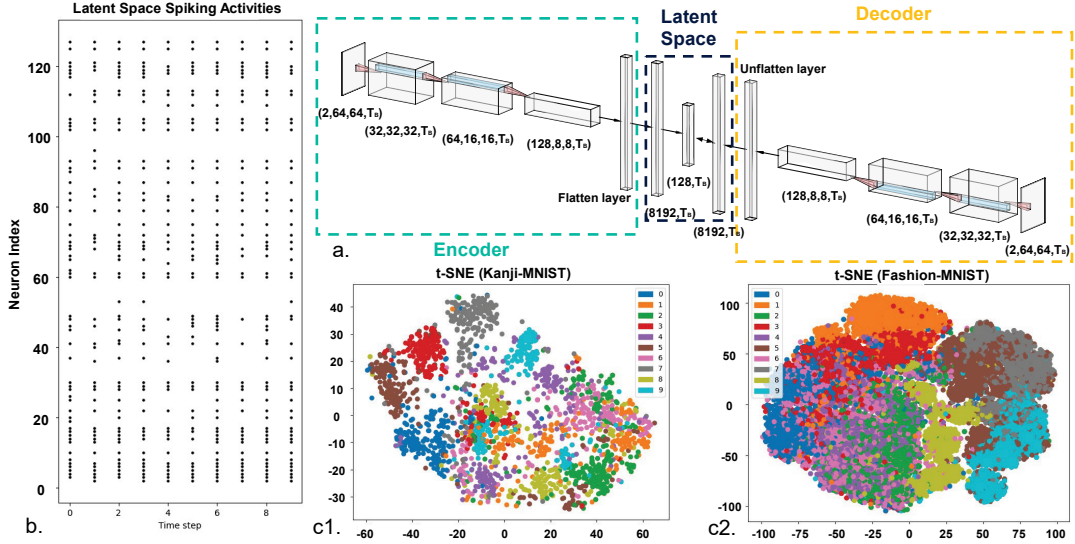

Fig 3. Structure and process flow of the deep SNN model, highlighting the spiking autoencoder (SAE) design and the role of the latent space within the model. (a) Illustrating the design of the SAE, here comprised of three blocks. The Encoder starts with an initial input matrix of [2,64,64] corresponding to 2 channels of positive and negative spikes (increase and decrease in optical intensity) across the 64×64 neurons (binned DVS pixels) in the x and y dimensions; three successive spiking convolutional neural layers then transform the spike data to size [128,8,8]. A flattening layer further converts this to a vector of size [8192]. The Latent Space operates within a reduced dimensionality space of [128] and is densely connected with two flat layers of both the encoder and decoder, each of dimension [8192]. The Decoder is designed as a reverse of the Encoder to produce a final output with dimensions identical to the initial input. Depending on the final layer's set threshold value, output spikes can either be accumulated as a continuous membrane potential or transformed back into final decoded spike trains. Note that the SAE design processes spikes at each time step with the final reconstruction resulting from multiple time steps within the same time block. (b) Tutorial view of the spiking activity of 128 model neurons in latent space over 10 time steps acquired from the same dynamic optical target, highlighting the similarity among spike trains post-encoding and reflecting the SAE's ability to capture and extract the target's essential features. (c) A t-SNE graphical analysis of spiking activities in the latent space (128 neurons, 10 time steps) for a set of 2000 Kanji-MNIST character targets. (d) Shows a t-SNE analysis of spiking activities in latent space (128 neurons, 10 time steps) for 30000 Fashion-MNIST image targets.

The implementation of the neuromorphic SNN algorithm described here offers several advantages. First, its ability to handle dynamic and complex data aligns perfectly with the intrinsic nature of the DVS data. Second, the SAE's effective filtering ability aids in the utilization of full spike trains thus minimizing data loss. Finally, the algorithm's capability to reconstruct *both* spatial and temporal information can bring added value to any number of imaging applications.

Still, the algorithm is not without limitations. While the algorithm can handle dynamic data, the complexity and computational demand is likely to rise with increasing complexity in the features of the targets of interest. Furthermore, the use of a spike similarity index as a loss function such as the van Rossum distance [27], described below, necessitates a spiking representation of the ground truth to be available. Often, the system design goal is to achieve an accurate and efficient latent spiking representation of data which is not intrinsically event-based, in which case the spike similarity index would not applicable. In this paper we show the benefits of the approach in reconstruction results below while acknowledging the need and opportunities for future exploration outside the scope of this paper.

**Training the Spiking Neural Network**

Training spiking neural networks (SNNs) poses its own challenges compared to traditional artificial neural networks. As most SNNs communicate using binarized all-or-nothing spikes, this leads to difficulties in applying gradient-based optimization methods like backpropagation. Although it is well-known that auto-differentiation frameworks are robust even in the case of globally non-differentiable neuron activation functions (see ReLU [28]), the issue arises due to the threshold spike generating function. When the membrane potential of a spiking neuron exceeds its threshold, it emits a discrete spike signal of 1, otherwise the value is 0. This behavior results in a gradient of either zero or infinity which prevents learning. This issue is commonly referred to as the "dead neuron problem" [29,30].

To overcome the challenge, we trained the deep convolutional spiking neural network using the method of Backpropagation Through Time (BPTT) by spikes with surrogate gradient descent. Though the network architecture consists of standard convolutional and dense connectivity, the spiking neuron models introduce temporal dynamics, making the network recurrent in nature. Specifically: (1) The output spikes are accumulated over multiple time steps before calculating the loss; (2) The membrane potential of the spiking neurons persists over time instead of being reset after each sample.

In brief, during the forward pass the input data consisted of spike trains that were fed sequentially through the network over multiple time steps. The output spikes were accumulated across all time steps before calculating the loss by comparing to target spikes. Depending on the outputs, we utilized Mean Square Error (MSE) and Spiking Train Similarity Distance (e.g. van Rossum distance) as the training objective for membrane potential and spike trains respectively. In Figs. 4 and 5, we employed the Mean Squared Error (MSE) as the loss function, while for Fig. 6, the Van Rossum distance was utilized as the loss metric.

Backpropagation was then applied to calculate gradients and update the weights to minimize error. We applied a smooth Arctangent surrogate gradient which has empirically proven effective for SNN training [31], to approximate the non-differentiable spike function. This allowed errors to flow backward through time and layers, providing gradients to update weights each synaptic weight parameter based on its contribution to the loss, as determined by BPTT, so as to reduce the loss. The surrogate gradient approach revived "dead" (silent) neurons by permitting gradients to flow even in the absence of spikes.

The network was trained across multiple epochs, gradually updating the weights to improve the performance. By applying BPTT, the network was able to learn temporal relationships in the input spike trains. The convolutional architecture extracted spatial features while the recurrent dynamics handled temporal dependencies. The architecture facilitated successful end-to-end training of the SNN by using the method of surrogate gradient descent.

**Case Studies of Image Reconstruction of Dynamic Targets Obscured by Dense Turbid Media**

Deploying the dual configuration of Fig. 1 we conducted a series of experiments in transmission and reflection/backscattering geometries, respectively, to assess the performance of our neuromorphic optical imaging approach. As the dynamical targets we chose various character and image sets from the MNIST data base, intending to span a wide range of shapes from simple to complex features so that a combination of these would likely cover a broad range of objects of interest in real world applications. The particular MNIST data sets were those for Kanji characters, Fashion images, and digits, respectively. Up to 60,000 images are in principle available for the SNN training with up to 10,000 as testing images. In some cases, we randomly chose a smaller number of samples from these datasets, as detailed below. Our rationale for deploying the Kanji characters was an observation that these contained features similar from biological objects such as cortical vasculature.

Figure 4 summarizes reconstruction results obtained in transmission geometry for randomly selected examples from the Kanji-MNIST and the Fashion-MNIST sets. Although presented here as static images, in each case the individual characters/images were actually "flashed" on and off by the DMD display at a rate of 1000 Hz to mimic a high-speed dynamical target. (see also the discussion in connection of Fig 6 for presentation of dynamical reconstruction as a video clip and online repository of Ref. 30). Here the full set of 60,000 characters was used for the Kanji set while 30,000 images were chosen from the Fashion Set. Using a standard (non-neuromorphic) workstation with a GPU (RTX3090, Nvidia), it typically took up to 8 hours to train for the full 60,000 sample dataset. The inference time, in turn, on a workstation using standard (non-neuromorphic) hardware is 38 ms in average for one target per reconstructed sample, a performance which is anticipated to be faster and more efficient when exploiting actual neuromorphic hardware. The figure shows a direct comparison between the 'ground truth' (software generated or experiment generated) data with the raw spiking data recorded by the DVS camera through the phantom (MFP =72) with the final SNN-generated reconstruction result. In these experiments the images were simple black-and-white without any grey scale because of the intrinsic on-off nature of the DMD projector. Even visually, the contrast between the raw DVS camera images (center column) and those generated by the SNN-algorithm offer clear evidence of the promise of the NCI approach. In particular, the algorithm successfully leveraged the use of the mean square error (MSE) as a loss function, with the final layer outputting membrane potentials from all neurons across all time steps of a time block. The membrane potentials were compared with the ground truth images, demonstrating a high degree of accuracy in the image reconstruction. In Fig 4a, the MSE is quantitatively reduced from 0.138 (between Input and Target) to 0.067 (between Reconstruction and Target). The structural similarity index (SSIM) has improved from 0.198 to 0.403. However, these quantified values may not fully capture the visual quality of the reconstruction as perceived by the human eye. This discrepancy is likely due to the semi-noisy background present in the reconstructions, in contrast to the clean black background of the target image. This can be further improved by employing advanced denoising algorithms, such as a U-net type structure, subsequent to the SNN output.

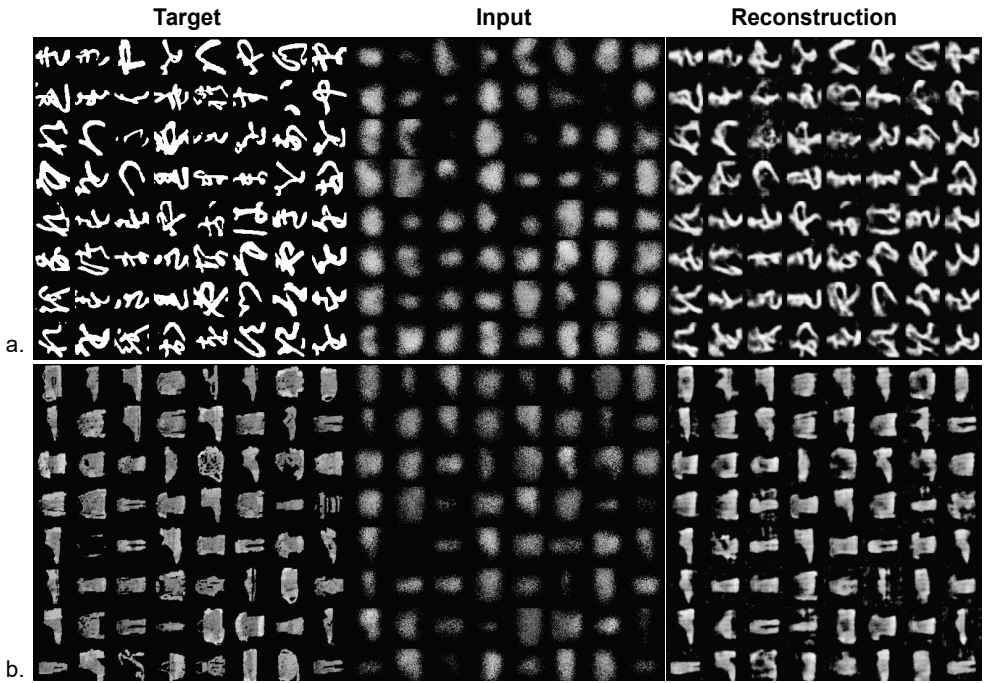

Fig 4. Results of reconstruction from transmission experiments in dense turbid media of MFP = 72 (using a DMD for the generation of target images as per Fig. 1). The three columns show a set of unobstructed DMD generated groundtruth images (left), the scattering-obscured raw images captured by the DVS camera (middle, generated by time integrated spikes), and results of SNN-based image reconstruction (right, generated by accumulated membrane potentials), respectively. (a) Using Kanji-MNIST characters as the dynamic target (total size of character set was 60000). Groundtruth is generated by software directly from the Kanji-MNIST dataset. Reconstruction and Target: SSIM = 0.403, MSE=0.067. Input and Target: SSIM = 0.198, MSE = 0.138. SSIM and MSE are averaged values from the illustrated 64 Kanji set. (b) Using Fashion-MNIST images as the dynamic targets (total image set was 30000). Groundtruth is acquired from experiments (time integrated DVS measured spikes across all time steps). Inputs to the SNN pipeline were formed from accumulation of spikes across all time steps with unobstructed target images representing ground truth.

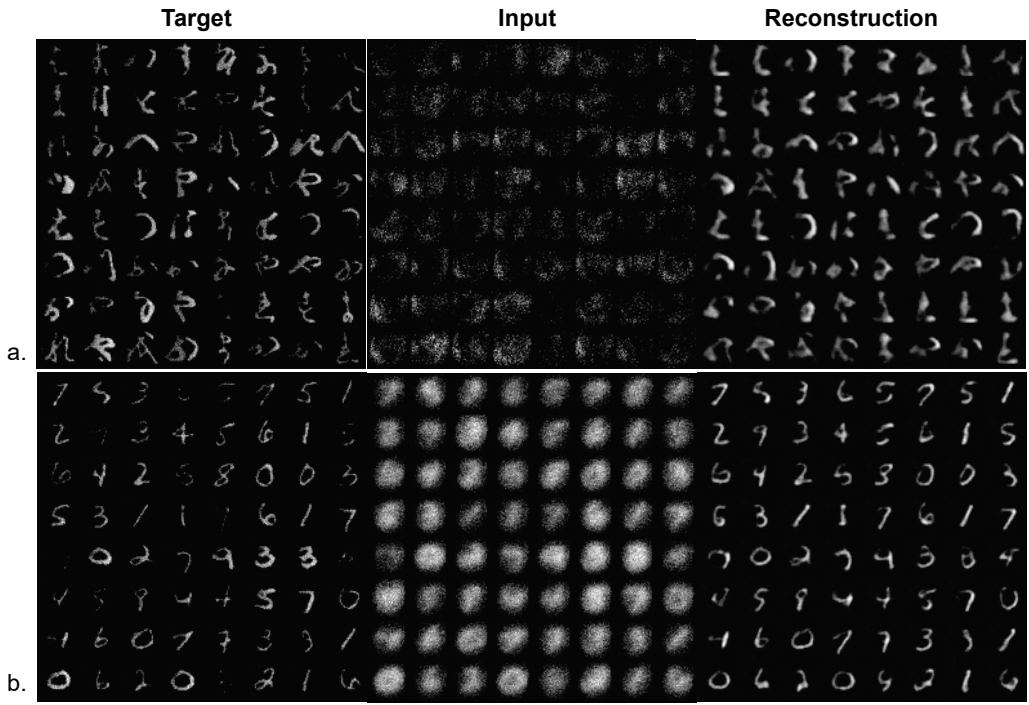

Fig 5. Examples of reconstruction results from reflection experiments in dense turbid media (MFP =60; target images generated by programming an e-ink display as in Fig.1). The three columns show a set of unobstructed groundtruth images from the E-ink display (left), scattering-obscured raw images captured by the DVS camera (middle), and results of SNN-based image reconstruction, respectively (right). The target and input images represent DVS measured spikes accumulated across all time steps. (a) The case where a set of Kanji-MNIST was used for the dynamic targets (total size of target set was 30000). Reconstruction and Target: SSIM = 0.326, MSE=0.0.021. Input and Target: SSIM = 0.094, MSE = 0.043. SSIM and MSE are averaged values from the illustrated 64 Kanji set. (b) MNIST digits as the dynamic targets (total size of the character set was 15000). The ground truths in (a) and (b) were both collected from benchtop experiments (time integrated DVS measured spikes across all time steps).

Figure 5 shows a result from the experiments in reflection geometry for Kanji and (numerical) digit characters, respectively. Again, severity of light scattering by the obscuring phantom renders the output from the DVS camera as being unrecognizable from the original input targets (here produced experimentally by the half part of E-ink display as the ground truth). However, when applying our custom NCI algorithm, the reconstructed images (right column) emerge as clear and unambiguous to the observer's eye. The results appear rather compelling in demonstrating that even with the challenge presented by the dense turbid medium, the combination of an event detecting camera and the SNN-based algorithm, with spikes as the basic unit of information, enable can led to the recovery of fully recognizable key features in the dynamical targets.

Finally, in order to demonstrate the capabilities of the dynamical image reconstruction in both spatial (2D images) and temporal dimensions (1D with time axis), Figure 6 shows an example of statistical representation of the spike-train map in both domains, here acquired from a reflection experiment for the digit "5". The map gives the reader visual context of how photons that have interacted with the dynamic target in the dense turbid media exhibit a distinctly different distribution from the case where scattering is absent. Additional reconstruction results including training and validation loss, animations of DVS raw measurements and reconstructed spike trains in temporal and spatial domain, please refer to our project's GitHub repository [32].

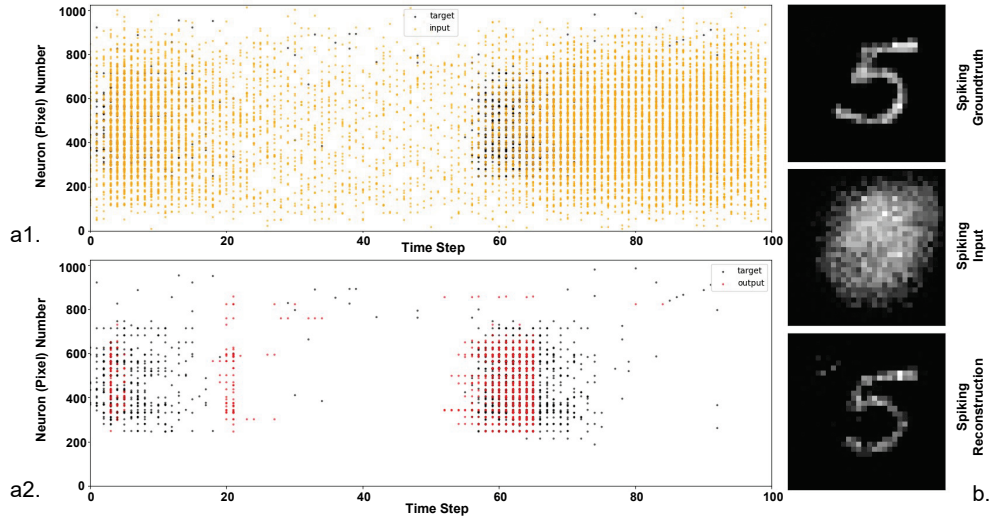

Fig 6. Illustrating spike train reconstruction in both temporal and spatial domain. The spike panels track the process of time-space reconstruction, here using as the dynamical target the digit "5" from the MNIST dataset as the example, the pattern flashed on-off by the E-ink display in reflection geometry . Left panels: Display of the spike raster plot of 1024 model neurons over 100 time steps, where the Y-axis encodes the spatial information leading to the target's image reconstruction while the X-axis conveys temporal information indicative of the target's dynamic behavior. (a1) illustrates spike trains from a DVS camera both with and without the scattering medium corresponding to a flash of digit "5" (the target spike trains revealed and marked as black dots). Majority of ost spikes with meaniful information are generated during the transient of the target appearing and disappearing whereas no spikes are contributed during the static phase of the display. The orange dots represent input spikes from photons which have interacted with the target and display a wider, more random distribution induced by light scattering (thereby destroying the original spatial and temporal information). (a2) Reconstructed output spike trains (red dots) juxtaposed on the detected target spike trains, showing that the SNN output (expressed here as spikes) can successfully reconstruct the target's shape (only specific neurons are actively spiking) and also restore its dynamic pattern (neurons firing at specific time steps). (b) Visual representations of the groundtruth (without scattering), the raw input to the SNN by the DVS camera, and the final image reconstruction, respectively. Video of the dynamical reconstruction can be found in Ref. 30.

## 3. Discussion

In this paper we have introduced a method to access dynamical targets obscured by turbid media by engineering a neuromorphic optical imaging system around a concept that mimics salient features of the human visual system. The system integrates an event-driven optical engineering subsystem with a neuromorphic decoding scheme whereby binary spikes carry and are deployed end-to-end as the sole currency unit of information. More specifically, spatio-temporal changes in light intensity which are collected by a DVS camera are transformed into spikes in real time akin to neural impulses in the retina, then forwarded to custom special purpose neuromorphic computational machinery for image reconstruction. In particular, we have developed an approach to deep learning using a SNN model and have demonstrated the ability of the algorithm for reconstruction of many different types MNIST-based images and characters. In the benchtop optical experiments of this paper we used dynamical targets which were spatially stationary but whose optical features were time-varying. However, our neuromorphic optical imaging approach can be readily extended to physically moving targets with case studies planned by us for future work.

There are a number of improvements which can be envisioned to build upon based on the results in this paper. On the optical engineering side, for example, the emerging availability of ultracompact semiconductor lasers such as vertical cavity laser (VCSEL) arrays offers the possibility for multichannel extension of what is essentially a single channel (single source) laser in this study [33,34]. Improving the sensitivity, contrast ratio, and noise performance of the event sensing cameras themselves would likewise enhance the capabilities of the approach. Similarly adding a time-of-flight (ToF) extension can add a significant further asset, provided that DVS-type cameras can be configured to include an ultrafast (< nsec) electronic shutter (as such, VCSEL arrays can readily reach multi-GHz speeds).

On the computational side, our choice of a spiking autoencoder (SAE) approach in the core of the SNN as well as the accompanying choice of the particular backpropagation method are but one of many possible ways to develop a deep SNN algorithmic strategy. There are likely many alternative application specific approaches and pathways given how neuromorphic computing is being continuously innovated as evidenced by the rapidly expanding literature [35].

As such, our study is a proof-of-concept demonstration of an entirely new approach to addressing an important problem of significant interest in pushing optical engineering and computational techniques towards the fundamental limits imposed by the inherently diffusive nature of photon propagation in turbid media [36,37]. With insight from biological vision and its ability specifically to identify dynamical targets, the work represents a synthesis of powerful camera technology (DVS detectors) with the opportunities proffered by neuromorphic computing (SNN model). The event sensing camera technology has matured and is now penetrating into applications such as machine vision in robotics and possesses a dynamical range and speed which can extend our approach to track fast moving targets in dense turbid media. In leveraging the inherent qualities of deep machine learning, here for the case of SNNs, our study demonstrates the promise these algorithm hold to reconstruct spatial and temporal information in difficult optical domains and paves the way to further development towards applications requiring dynamical object reconstruction in real time. This result offers considerable potential for optically imaging complex phenomena that are shrouded in complex scattering media, including orbitally-monitored wildfires, vehicles obscured in fog, and in-vivo organ monitoring. The technique should also be applicable to use in microscopy of biological media where recent progress has been made in computational imaging [38] as well as adding to the photonic toolkit useful for neuroimaging [39]. Another potential direction is the use of the neuromorphic computational approach in advancing non-line of sight imaging methods [40]. With the emphasis on using binary spikes as the unit of information, the approach of this paper can be extended to include low-error rate wireless transmission of data prior to decoding [41]. Finally, given the emergence of low-power, low-latency, compact portable neuromorphic hardware such as Intel's Loihi 2[1], the scheme presented in this paper could be particularly attractive for those optical sensor applications in the field where mobility is of the essence.

4. Materials and Methods

**Phantom preparation**

We strictly follow the methods and materials described in references[16,42] to fabricate the optical phantoms in experiments. The primary substance for creating optical phantoms is SiliGlass, a transparent silicone rubber procured from MBFibreglass (PlatSil SiliGlass), which has two components: part A (base material) and part B (hardener). To achieve specific optical properties, a black silicone pigment known as Polycraft Black Silicone Pigment is used as an absorber (diluted in part A at a ratio of 2272:1), and silica microspheres (440345, Sigma-Aldrich) are employed for scattering. Phantom preparation involves mixing part A of SiliGlass with the diluted absorber for absorption and part B with the disaggregated microspheres for scattering. Both mixtures are stirred magnetically and ultrasonicated for 15 minutes each. After combining at a controlled temperature of 21°C, they are stirred for uniformity and poured into a 3D-printed mold. The resulting phantoms, at different thickness, are designed to represent specific optical properties in the paper.

**DVS configuration and synchronization**

We utilized the Prophesee Evaluation Kit 3.0 – VGA as our DVS camera. This camera has a sensor resolution of 640 x 480, with each pixel measuring 15 x 15 μm. It boasts a latency of 220 μs and a dynamic range exceeding 120dB. By default, the nominal contrast threshold for detecting a new event is set to 25%. However, adjusting the sensor's bias can reduce this threshold to 12%[43]. To enhance photon collection, a C/CS mount lens system is mounted above the camera sensor. During our experiments, the lens aperture was maintained at f/2.8 to optimize light entry to the sensor.

The DVS camera is equipped with synchronization input and output capabilities, enabling the detection of external signals for coordinating between multiple devices. Here, a rising signal edge is interpreted as a start signal, while a falling edge signifies a stop signal. Throughout our experiments, the DVS camera remained operational, ensuring uninterrupted event capture. As part of our display procedure for a target, both the e-ink display and DMD (based on the experimental type) sent a 5V TTL square wave signal. This signal's rising edge preceded the display onset, and its falling edge marked the display conclusion. The DVS system recognized and timestamped these synchronization signals alongside events, offering a microsecond resolution. Once all targets were displayed, the accumulated spike trains (events) were retrieved and partitioned into distinct recordings, according to the identified trigger signals. Subsequent to this, each recording underwent preprocessing to filter out noise events and were tagged with the corresponding target identity, readying them for the subsequent training and testing phases.

**Data preprocessing**

Before spike trains are fed into the deep SNN, they often necessitate a preprocessing step. The initial step involves selecting a region of interest (ROI). Only spikes from this defined ROI are permitted to proceed to the subsequent process of noise filtering. There are multiple algorithms available for this stage:

(1) Activity-Noise Filter: This filter permits events if a similar event has occurred within a specific time frame in the past, in proximity to its coordinates.

(2) Spatio-Temporal-Contrast Filter: This method leverages the exponential response of a pixel to light changes in order to filter out inaccurate detections and trails. For an event to be passed on, it must be preceded by another event within a set time frame. This ensures robust spatio-temporal contrast detection.

(3) AntiFlicker Filter: Designed to eliminate flickering events within a given frequency range, this filter is commonly employed to sift out ambient light noise exhibiting a distinct frequency.

Of these, we found the Activity-Noise Filter to be the most practically beneficial in effectively filtering out noise events during our experiments.

Following noise filtering, the spike trains undergo binning in both spatial and temporal domains. Given that our existing network interprets each DVS camera pixel as an input neuron, it's impractical to utilize the full 307,200 pixels as input neurons. In our preprocessing phase, we binned the ROI pixels into either 32x32 (yielding 1,024 input neurons) or 64x64 (yielding 4,096 input neurons) dimensions.

A similar challenge arises in the temporal domain. Training with spiking events is computationally demanding, as it involves loading spikes through iterative time steps. Given the high-speed, microsecond-resolution capabilities of a DVS camera, even a brief 1 ms recording (representative of a target in our DMD transmission experiment) contains 1,000 time steps to process. Consequently, condensing spikes into smaller time bins becomes imperative, albeit at the expense of temporal resolution. This situation presents a trade-off between training/inference speed and the temporal resolution of target dynamics. In our experiment, we empirically opted to bin spikes into 50 time steps.

**SNN Training and testing environments**

We set up our training and testing environment with the open-source Lava framework for neuromorphic computing [44] and snnTorch library [45].

**Acknowledgement:** We thank Jordan Watts, Jihun Lee, Mike Davies, Sumit Bam Shrestha, Garrick Orchard, Pedro Felzenswalb, and Quang Zhang for their insight and advice. Research was supported by a grant from Intel Labs, NIH Award 1S10OD025181, and a private gift.

**Author contributions:** N.Z. and A.N. conceived the concept of the work. N.Z. built the system, designed and performed experiments, and processed the data. N.Z. and T.S. developed the computational model. N.Z. and A.N. wrote the manuscript draft. All authors were involved in discussions and contributed to the manuscript editing.

**Conflict of interest:** The authors declare no conflict of interests.

**Data availability:** Data underlying the results presented in this paper is available in the provided GitHub repository. More information may be obtained from the authors upon reasonable request.